\documentclass[11pt,a4paper]{article}
\usepackage[hyperref]{emnlp-ijcnlp-2019}
\usepackage{times}
\usepackage{latexsym}
\usepackage{amsmath}
\usepackage{amssymb}
\usepackage{graphicx}
\usepackage{framed}
\usepackage[title]{appendix}
\usepackage{url}

\aclfinalcopy 

\usepackage{array}
\newcolumntype{L}[1]{>{\raggedright\let\newline\\\arraybackslash\hspace{0pt}}m{#1}}
\newcolumntype{C}[1]{>{\centering\let\newline\\\arraybackslash\hspace{0pt}}m{#1}}
\newcolumntype{R}[1]{>{\raggedleft\let\newline\\\arraybackslash\hspace{0pt}}m{#1}}

\title{Cost-Sensitive BERT for Generalisable \\ Sentence Classification with Imbalanced Data}

\author{
Harish Tayyar Madabushi\textsuperscript{1} \and 
Elena Kochkina\textsuperscript{2,3} \and 
Michael Castelle\textsuperscript{2,3} 
\\
\\
\textsuperscript{1} University of Birmingham, UK\\
\tt{H.TayyarMadabushi.1@bham.ac.uk}\\
\\
\textsuperscript{2}University of Warwick, UK\\
\tt{(E.Kochkina,M.Castelle.1)@warwick.ac.uk}
\\
\textsuperscript{3}Alan Turing Institute, UK\\
}
  
\date{}

\begin{document}
\maketitle
\begin{abstract}
 
 The automatic identification of propaganda has gained significance in recent years due to technological and social changes in the way news is generated and consumed. That this task can be addressed effectively using BERT, a powerful new architecture which can be fine-tuned for text classification tasks, is not surprising. However, propaganda detection, like other tasks that deal with news documents and other forms of decontextualized social communication  (e.g. sentiment analysis), inherently deals with data whose categories are simultaneously imbalanced and dissimilar. We show that BERT, while capable of handling imbalanced classes with no additional data augmentation, does not generalise well when the training and test data are sufficiently dissimilar (as is often the case with news sources, whose topics evolve over time). We show how to address this problem by providing a statistical measure of similarity between datasets and a method of incorporating cost-weighting into BERT when the training and test sets are dissimilar. We test these methods on the Propaganda Techniques Corpus (PTC) and achieve the second highest score on sentence-level propaganda classification.
\end{abstract}

\section{Introduction}

The challenges of imbalanced classification---in which the proportion of elements in each class for a classification task significantly differ---and of the ability to generalise on dissimilar data have remained important problems in Natural Language Processing (NLP) and Machine Learning in general. Popular NLP tasks including sentiment analysis, propaganda detection, and event extraction from social media are all examples of imbalanced classification problems. In each case the number of elements in one of the classes (e.g. negative sentiment, propagandistic content, or specific events discussed on social media, respectively) is significantly lower than the number of elements in the other classes. 

The recently introduced BERT language model for transfer learning~\cite{devlin2018bert} uses a deep bidirectional transformer architecture to produce pre-trained context-dependent embeddings. It has proven to be powerful in solving many NLP tasks and, as we find, also appears to handle imbalanced classification well, thus removing the need to use standard methods of data augmentation to mitigate this problem (see Section \ref{section:lit-review-data-aug} for related work and Section \ref{section:classimbalance} for analysis).

BERT is credited with the ability to adapt to many tasks and data with very little training~\cite{devlin2018bert}. However, we show that BERT fails to perform well when the training and test data are significantly dissimilar, as is the case with several tasks that deal with social and news data. In these cases, the training data is necessarily a subset of past data, while the model is likely to be used on future data which deals with different topics. This work addresses this problem by incorporating cost-sensitivity (Section \ref{section:costsensitive}) into BERT. 

We test these methods by participating in the Shared Task on Fine-Grained Propaganda Detection for the 2nd Workshop on NLP for Internet Freedom, for which we achieve the second rank on sentence-level classification of propaganda, confirming the importance of cost-sensitivity when the training and test sets are dissimilar.

\subsection{Detecting Propaganda}

The term `propaganda' derives from {\it propagare} in post-classical Latin, as in ``propagation of the faith" \cite{auerbach_oxford_2014}, and thus has from the beginning been associated with an intentional and potentially multicast communication; only later did it become a pejorative term. It was pragmatically defined in the World War II era as 
``the expression of an opinion or an action by individuals or groups deliberately designed to influence the opinions or the actions of other individuals or groups with reference to predetermined ends" \cite{institute1937detect}. 

For the philosopher and sociologist Jacques Ellul, however, in a society with mass communication, propaganda is {\it inevitable} and thus it is necessary to become more {\it aware} of it \cite{ellul_propaganda_1973}; but whether or not to classify a given strip of text as propaganda depends not just on its content but on its {\it use} on the part of both addressers and addressees \cite[6]{auerbach_oxford_2014}, and this fact makes the automated detection of propaganda intrinsically challenging.

Despite this difficulty, interest in automatically detecting misinformation and/or propaganda has gained significance due to the exponential growth in online sources of information combined with the speed with which information is shared today. The sheer volume of social interactions makes it impossible to manually check the veracity of all information being shared. Automation thus remains a potentially viable method of ensuring that we continue to enjoy the benefits of a connected world without the spread of misinformation through either ignorance or malicious intent. 

In the task introduced by \citet{EMNLP19DaSanMartino}, we are provided with articles tagged as propaganda at the sentence and fragment (or span) level and are tasked with making predictions on a development set followed by a final held-out test set. We note this gives us access to the articles in the development and test sets but not their labels. 

We participated in this task under the team name {\it ProperGander} and were placed 2\textsuperscript{nd} on the sentence level classification task where we make use of our methods of incorporating cost-sensitivity into BERT. We also participated in the fragment level task and were placed 7\textsuperscript{th}. The significant contributions of this work are: 
\begin{itemize}
    \setlength\itemsep{0.05em}
    \item We show that common (`easy') methods of data augmentation for dealing with class imbalance do not improve base BERT performance.
    \item We provide a statistical method of establishing the similarity of datasets. 
    \item We incorporate cost-sensitivity into BERT to enable models to adapt to dissimilar datasets.    
    \item We release all our program code on GitHub and Google Colaboratory\footnote{\url{http://www.harishmadabushi.com/research/propaganda-detection/}}, so that other researchers can benefit from this work.
\end{itemize}
 
\section{Related work}

\subsection{Propaganda detection}

Most of the existing works on propaganda detection focus on identifying propaganda at the news article level, or even at the news outlet level with the assumption that each of the articles of the suspected propagandistic outlet are propaganda \cite{rashkin2017truth,barron2019proppy}. 

Here we study two tasks that are more fine-grained, specifically propaganda detection at the sentence and phrase (fragment) levels \cite{EMNLP19DaSanMartino}. This fine-grained setup aims to train models that identify linguistic propaganda techniques rather than distinguishing between the article source styles. 

\citeauthor{EMNLP19DaSanMartino} \shortcite{EMNLP19DaSanMartino} were the first to propose this problem setup and release it as a shared task.\footnote{\url{https://propaganda.qcri.org/nlp4if-shared-task/}} Along with the released dataset, ~\citet{EMNLP19DaSanMartino}  proposed a multi-granularity neural network, which uses the deep bidirectional transformer architecture known as BERT, which features pre-trained context-dependent embeddings \cite{devlin2018bert}. Their system takes a joint learning approach to the sentence- and phrase-level tasks, concatenating the output representation of the less granular (sentence-level) task with the more fine-grained task using learned weights.  

In this work we also take the BERT model as the basis of our approach and focus on the class imbalance as well as the lack of similarity between training and test data inherent to the task.

\subsection{Class imbalance}

A common issue for many Natural Language Processing (NLP) classification tasks is {\it class imbalance}, the situation where one of the class categories comprises a significantly larger proportion of the dataset than the other classes. It is especially prominent in real-world datasets and complicates classification when the identification of the minority class is of specific importance.

Models trained on the basis of minimising errors for imbalanced datasets tend to 
more frequently predict the majority class; achieving high accuracy in such cases can be misleading. Because of this, the macro-averaged F-score, chosen for this competition, is a more suitable metric as it weights the performance on each class equally.

As class imbalance is a widespread issue, multiple techniques have been developed that help alleviate it \cite{buda2018systematic,haixiang2017learning}, by either adjusting the model (e.g. changing the performance metric) or changing the data (e.g. oversampling the minority class or undersampling the majority class). 

\subsubsection{Cost-sensitive learning} 
\label{section:costsensitiveintro}
Cost-sensitive classification can be used when the ``cost'' of mislabelling one class is higher than that of mislabelling other classes~\cite{elkan2001foundations,kukar1998cost}. For example, the real cost to a bank of miscategorising a large fraudulent transaction as authentic is potentially higher than miscategorising (perhaps only temporarily) a valid transaction as fraudulent. Cost-sensitive learning tackles the issue of class imbalance by changing the cost function of the model such that misclassification of training examples from the minority class carries more weight and is thus more `expensive'. This is achieved by simply multiplying the loss of each example by a certain factor. This cost-sensitive learning technique takes misclassification costs into account during 
model training, and does not modify the imbalanced data distribution directly.

\subsubsection{Data augmentation} \label{section:lit-review-data-aug}
Common methods that tackle the problem of class imbalance by modifying the data to create balanced datasets are undersampling and oversampling. Undersampling randomly removes instances from the majority class and is only suitable for problems with an abundance of data. Oversampling means creating more minority class instances to match the size of the majority class. 
Oversampling methods range from simple 
random oversampling, i.e. repeating the training procedure on instances from the minority class, chosen at random, to the more complex, which involves constructing synthetic minority-class samples. Random oversampling is similar to cost-sensitive learning as repeating the sample several times makes the cost of its mis-classification grow proportionally. Kolomiyets et al.~\shortcite{kolomiyets2011model}, Zhang et al.~\shortcite{zhang2015character}, and Wang and Yang ~\shortcite{wang2015s} perform data augmentation using synonym replacement, i.e. replacing random words in sentences with their synonyms or nearest-neighbor embeddings, and show its effectiveness on multiple tasks and datasets.  
Wei et al.~\shortcite{wei2019eda} provide a great overview of `easy' data augmentation (EDA) techniques for NLP, including synonym replacement as described above, and random deletion, i.e. removing words in the sentence at random with pre-defined probability. They show the effectiveness of EDA across five text classification tasks. However, they mention that EDA may not lead to substantial improvements when using pre-trained models.  
In this work we test this claim by comparing performance gains of using cost-sensitive learning versus two data augmentation methods, synonym replacement and random deletion, with a pre-trained BERT model. 

More complex augmentation methods include back-translation \cite{sennrich2015improving}, translational data augmentation \cite{fadaee2017data}, and noising \cite{xie2017data}, but these are out of the scope of this study.

\section{Dataset}

The Propaganda Techniques Corpus (PTC) dataset for the 2019 Shared Task on Fine-Grained Propaganda consists of 
a {\it training set} of 350 news articles, consisting of just over 16,965 total sentences, in which specifically propagandistic fragments have been manually spotted and labelled by experts. This is accompanied by a {\it development set} (or {\it dev set}) of 61 articles with 2,235 total sentences, whose labels are maintained by the shared task organisers; and two months after the release of this data, the organisers released a {\it test set} of 86 articles and 3,526 total sentences. In the training set, 4,720 ($\sim 28\%$) of the sentences have been assessed as containing propaganda, with 12,245 sentences ($\sim  72 \%$) as non-propaganda, demonstrating a clear class imbalance.

In the binary {\it sentence-level classification} (SLC) task, a model is trained to detect whether each and every sentence is either 'propaganda' or 'non-propaganda'; in the more challenging {\it field-level classification} (FLC) task, a model is trained to detect one of 18 possible propaganda technique types in spans of characters {\it within} sentences. These propaganda types are listed in \citet{EMNLP19DaSanMartino} and range from those which might be recognisable at the lexical level (e.g. {\sc Name\_Calling}, {\sc Repetition}), and those which would likely need to incorporate semantic understanding ({\sc Red\_Herring}, {\sc Straw\_Man}).\footnote{\url{https://propaganda.qcri.org/annotations/} includes a flowchart instructing annotators to discover and isolate these 18 propaganda categories.}

For several example sentences from a sample document annotated with fragment-level classifications (FLC) (Figure \ref{fig:kavanaugh}). The corresponding sentence-level classification (SLC) labels would indicate that sentences 3, 4, and 7 are 'propaganda' while the the other sentences are `non-propaganda'.

\begin{figure*}
 \footnotesize
 \begin{tabular}{|C{2cm}L{13cm}|}
 \hline
&\\ [-1.5ex]
   \textbf{Sentence 1:} & The Senate Judiciary Committee voted 11-10 along party lines to advance the nomination of Judge Brett Kavanaugh out of committee to the Senate floor for a vote. \\ [1ex]
   \textbf{Sentence 2:} & Of course, RINO Senator Jeff Flake (R-AZ) wanted to side with Senate Democrats in pushing for a FBI investigation into unsubstantiated allegations against Kavanaugh. \\ [1ex]
   \textbf{Sentence 3:} & Outgoing Flake, and $<${\sc  Loaded\_Language}$>$ {\bf good riddance} $<${\sc /Loaded\_Language}$>$, said that he sided with his colleagues in having a "limited time and scope" investigation by the FBI into the allegations against Kavanaugh. \\ [1ex]
   \textbf{Sentence 4:} & ``$<${\sc Flag-Waving}$>$
\ {\bf This country is being ripped apart here, and we've got to make sure we do due diligence}$<${\sc /Flag-Waving}$>$,'' Flake said. \\ [1ex]
   \textbf{Sentence 5:} & He added that he would be more "comfortable" with an FBI investigation. \\ [1ex]
   \textbf{Sentence 6:} & Comfort?\\ [1ex]
   \textbf{Sentence 7:} & $<${\sc Whataboutism}$>${\bf What about Judge Kavanaugh's comfort in being put through the ringer without a shred of evidence, Senator Flake}$<${\sc /Whataboutism}$>$? \\[1ex]
   \hline
 \end{tabular}
 \caption{Excerpt of an example (truncated) news document with three separate field-level classification (FLC) tags, for \sc Loaded Language, Flag-Waving, and Whataboutism.}
 \label{fig:kavanaugh}
\vspace{-1.5em}
\end{figure*}

\subsection{Data Distribution}
One of the most interesting aspects of the data provided for this task is the notable difference between the training and the development/test sets. We emphasise that this difference is realistic and reflective of real world news data, in which major stories are often accompanied by the introduction of new terms, names, and even phrases. This is because the training data is a subset of past data while the model is  to be used on future data which deals with different newsworthy topics.

We demonstrate this difference statistically by using a method for finding the similarity of corpora suggested by ~\citet{jbp:/content/journals/10.1075/ijcl.6.1.05kil}. We use the Wilcoxon signed-rank test~\cite{wilcoxon1945individual} which compares the frequency counts of randomly sampled elements from different datasets to determine if those datasets have a statistically similar distribution of elements. 

We implement this as follows. For each of the training, development and test sets, we extract all words (retaining the repeats) while ignoring a set of stopwords (identified through the Python Natural Language Toolkit). We then extract 10,000 samples (with replacements) for various pairs of these datasets (training, development, and test sets along with splits of each of these datasets). Finally, we use comparative word frequencies from the two sets to calculate the p-value using the Wilcoxon signed-rank test. Table \ref{table:mwut} provides the minimum and maximum p-values and their interpretations for ten such runs of each pair reported. 
\begin{table}[ht]
\begin{center}
\scriptsize	
\begin{tabular}{|L{1.cm}L{1.cm}C{1cm}C{1cm}C{1.6cm}|}
\hline
\textbf{Set 1} & \textbf{Set 2} & \textbf{p-value \ (min)} & \textbf{p-value \ (max)} & \textbf{\%} \ \textbf{Similar Tests} \\  [0.5ex]
\hline 
50\% Train   &	50\% Train   &	2.38E-01    &	9.11E-01    &	100 \\  [0.5ex]
50\% Dev     &	50\% Dev     &	5.55E-01    &	9.96E-01    &	100 \\  [0.5ex]
50\% Test	&   50\% Test    &	6.21E-01    &	8.88E-01    &	100 \\  [0.5ex]
25\% Dev     &	75\% Dev     &	1.46E-01    &	5.72E-01    &	100 \\  [0.5ex]
25\% Test    &   75\% Test	&   3.70E-02    &	7.55E-01    &	90  \\  [0.5ex]
25\% Train    &   75\% Train    & 	9.08E-02    &	9.66E-01    &	100 \\  [0.5ex]
Train           &	Dev             &	2.05E-09    &	4.33E-05    &	0   \\  [0.5ex]
Train           &	Test            &	8.37E-23    &	1.18E-14    &	0   \\  [0.5ex]
Dev	            &   Test            &	2.72E-04    &	2.11E-02    &	0   \\  [0.5ex]

\hline
\end{tabular}
\caption{\label{table:mwut}p-values representing the similarity between (parts of) the train, test and development sets.}
\end{center}
\vspace{-5mm}
\end{table}
With p-value less than 0.05, we show that the train, development and test sets are self-similar and also significantly different from each other. In measuring self-similarity, we split each dataset after shuffling all sentences. While this comparison is made at the sentence level (as opposed to the article level), it is consistent with the granularity used for propaganda detection, which is also at the sentence level. We also perform measurements of self similarity after splitting the data at the article level and find that the conclusions of similarity between the sets hold with a p-value threshold of 0.001, where p-values for similarity between the training and dev/test sets are orders of magnitude lower compared to self-similarity. Since we use random sampling we run this test 10 times and present the both the maximum and minimum p-values. We include the similarity between 25\% of a dataset and the remaining 75\% of that set because that is the train/test ratio we use in our experiments, further described in our methodology (Section \ref{section:methodology}). 

This analysis shows that while all splits of each of the datasets are statistically similar, the training set (and the split of the training set that we use for experimentation) are significantly different from the development and test sets. While our analysis does show that the development and the test sets are dissimilar, we note (based on the p-values) that they are significantly more similar to each other than they are to the training set.

\section{Methodology} \label{section:methodology}
We were provided with two tasks: (1) propaganda fragment-level identification (FLC) and (2)  propagandistic sentence-level identification (SLC). 
While we develop systems for both tasks, our main focus is toward the latter. Given the differences between the training, development, and test sets, we focus on methods for generalising our models. We  note that propaganda identification is, in general, an imbalanced binary classification problem as most sentences are not propagandistic. 

Due to the non-deterministic nature of fast GPU computations, we run each of our models three times and report the average of these three runs through the rest of this section. When picking the model to use for our final submission, we pick the model that performs best on the development set. 

When testing our models, we split the labelled training data into two non-overlapping parts: the first one, consisting of 75\% of the training data is used to train models, whereas the  other is used to test the effectiveness of the models. All models are trained and tested on the same split to ensure comparability. Similarly, to ensure that our models remain comparable, we continue to train on the same 75\% of the training set even when testing on the development set. 

Once the best model is found using these methods, we train that model on all of the training data available before then submitting the results on the development set to the leaderboard. These results are detailed in the section describing our results (Section \ref{section:results}).

\subsection{Class Imbalance in Sentence Level Classification} \label{section:classimbalance}

The sentence level classification task is an imbalanced binary classification problem that we address using BERT~\cite{devlin2018bert}. We use BERT\textsubscript{BASE}, uncased, which consists of 12 self-attention layers, and returns a 768-dimension vector that representation a sentence. So as to make use of BERT for sentence classification, we include a fully connected layer on top of the BERT self-attention layers, which classifies the sentence embedding provided by BERT into the two classes of interest (propaganda or non-propaganda). 

 We attempt to exploit various data augmentation techniques to address the problem of class imbalance. Table \ref{table:augresults} shows the results of our experiments for different data augmentation techniques when, after shuffling the training data, we train the model on 75\% of the training data and test it on the remaining 25\% of the training data and the development data. 

\begin{table}[ht]
\begin{center}
\footnotesize
\begin{tabular}{|L{2cm}C{2cm}C{2cm}|}
\hline
\textbf{Augmentation \ Technique} &    \textbf{f1-score on \ 25\% of Train}   &  \textbf{f1-score on \ Dev} \\[0.5ex]
\hline
None                &   0.7954      &   0.5803 \\[0.5ex]
Synonym Insertion   &   0.7889      &   0.5833 \\[0.5ex]
Dropping Words      &   0.7791      &   0.5445 \\[0.5ex]
Over Sampling       &   0.7843      &   0.6276 \\[0.5ex]
\hline
\end{tabular}
\caption{\label{table:augresults}F1 scores on an unseen (not used for training) part of the training set and the development set on BERT using different augmentation techniques. }
\end{center}
\vspace{-4mm}
\end{table}

We observe that BERT \emph{without} augmentation consistently outperforms BERT with augmentation in the experiments when the model is trained on 75\% of the training data and evaluated on the rest, i.e trained and evaluated on \emph{similar data}, coming from the same distribution.  This is consistent with observations by Wei et al.~\shortcite{wei2019eda} that contextual word embeddings do not gain from data augmentation. The fact that we shuffle the training data prior to splitting it into training and testing subsets could imply that the model is learning to associate topic words, such as `Mueller', as propaganda.
However, when we perform model evaluation using the development set, which is dissimilar to the training, we observe that synonym insertion and word dropping techniques also do not bring performance gains, while random oversampling increases performance over base BERT by 4\%. Synonym insertion provides results very similar to base BERT, while random deletion harms model performance producing lower scores.
We believe that this could be attributed to the fact that synonym insertion and random word dropping involve the introduction of noise to the data, while oversampling does not. As we are working with natural language data, this type of noise can in fact change the meaning of the sentence. Oversampling on the other hand purely increases the importance of the minority class by repeating training on the unchanged instances. 

So as to better understand the aspects of oversampling that contribute to these gains, we perform a class-wise performance analysis of BERT with/without oversampling. The results of these experiments (Table \ref{table:oversampling}) show that oversampling increases the overall recall while maintaining precision. This is achieved by significantly improving the recall of the minority class (propaganda) at the cost of the recall of the majority class. 

\begin{table}[ht]
\begin{center}
\footnotesize
\begin{tabular}{|L{3cm}L{1cm}L{1cm}|}
\hline
\textbf{} &    \textbf{OS}   &  \textbf{No OS} \\[0.5ex]
\hline
precision	              & 0.7967          &       0.7933   \\[0.5ex]
recall	                  & 0.7767          &       0.8000      \\[0.5ex]
f1-score	              & 0.7843          &       0.7954   \\[0.5ex]
&&\\[-1.5ex]
Non-Propaganda precision  & 0.8733          &       0.8467   \\[0.5ex]
Non-Propaganda recall	  & 0.8100            &       0.8900     \\[0.5ex]
Non-Propaganda F1	      & 0.8433          &       0.8667   \\[0.5ex]
&&\\[-1.5ex]
Propaganda precision	  & 0.5800            &       0.6600     \\[0.5ex]
Propaganda recall	      & 0.6933          &       0.5533   \\[0.5ex]
Propaganda F1             & 0.6300            &       0.6033   \\[0.5ex]
\hline  
\end{tabular}
\caption{\label{table:oversampling}Class-wise precision and recall with and without oversampling (OS) achieved on unseen part of the training set.}
\end{center}
\vspace{-4mm}
\end{table}

So far we have been able to establish that \textbf{a)} the training and test sets are dissimilar, thus requiring us to generalise our model, \textbf{b)} oversampling provides a method of generalisation, and \textbf{c)} oversampling does this while maintaining recall on the minority (and thus more interesting) class. 

Given this we explore alternative methods of increasing minority class recall without a significant drop in precision. One such method is cost-sensitive classification, which differs from random oversampling in that it provides a more continuous-valued and consistent method of weighting samples of imbalanced training data; for example, random oversampling will inevitably emphasise some training instances at the expense of others. We detail our methods of using cost-sensitive classification in the next section. Further experiments with oversampling might have provided insights into the relationships between these methods, which we leave for future exploration.

\subsection{Cost-sensitive Classification}\label{section:costsensitive}

As discussed in Section \ref{section:costsensitiveintro},
 cost-sensitive classification can be performed by weighting the cost function. We increase the weight of incorrectly labelling a propagandistic sentence by altering the cost function of the training of the final fully connected layer of our model previously described in Section \ref{section:classimbalance}. We make these changes through the use of PyTorch~\cite{paszke2017automatic} which calculates the cross-entropy loss for a single prediction $x$, an array where the $j^{th}$ element represents the models prediction for class $j$, labelled with the class $class$ as given by Equation \ref{equation:crossent}. 

\begin{equation} \label{equation:crossent}
\begin{gathered}
\text{loss}(x, class) = -\log\left(\frac{\exp(x[class])}{\sum_j \exp(x[j])}\right) \\
                       = -x[class] + \log\left(\sum_j \exp(x[j])\right)
\end{gathered}
\end{equation}

The cross-entropy loss given in Equation \ref{equation:crossent} is modified to accommodate an array $weight$, the $i^{th}$ element of which represents the weight of the $i^{th}$ class, as described in Equation \ref{equation:crossent-weight}. 

\begin{equation} \label{equation:crossent-weight}
\begin{gathered}
\text{loss}(x, class) = weight[class] \Theta \\
\text{where, }
\Theta = -x[class] + \log\left(\sum_j \exp(x[j])\right)
\end{gathered}
\end{equation}

Intuitively, we increase the cost of getting the classification of an ``important'' class wrong and corresponding decrees the cost of getting a less important class wrong. In our case, we increase the cost of mislabelling the minority class which is ``propaganda'' (as opposed to ``non-propaganda''). 

We expect the effect of this to be similar to that of oversampling, in that it is likely to enable us to increase the recall of the minority class thus resulting in the decrease in recall of the overall model while maintaining high precision. We reiterate that this specific change to a model results in increasing the model's ability to better identify elements belonging to the minority class in \emph{dissimilar} datasets when using BERT. 

\begin{figure}
    \centering
    \includegraphics[scale=0.5]{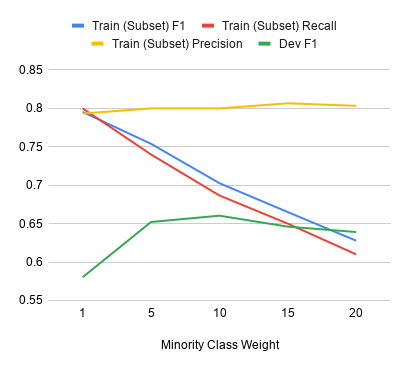}
    \caption{The impact of modifying the minority class weights on the performance on similar (subset of training set) and dissimilar (development) datasets. The method of increasing minority class weights is able to push the model towards generalisation while maintaining precision.}
    \label{fig:chart}
    \vspace{-4mm}
\end{figure}

We explore the validity of this by performing several experiments with different weights assigned to the minority class. 
We note that in our experiments use significantly higher weights than the weights proportional to class frequencies in the training data, that are common in literature \cite{ling24cost}. Rather than directly using the class proportions of the training set, we show that tuning weights based on performance on the development set is more beneficial. Figure \ref{fig:chart} shows the results of these experiments wherein we are able to maintain the precision on the subset of the training set used for testing while reducing its recall and thus generalising the model. The fact that the model is generalising on a dissimilar dataset is confirmed by the increase in the development set F1 score. We note that the gains are not infinite and that a balance must be struck based on the amount of generalisation and the corresponding loss in accuracy. The exact weight to use for the best transfer of classification accuracy is related to the dissimilarity of that other dataset and hence is to be obtained experimentally through hyperparameter search. Our experiments showed that a value of 4 is best suited for this task. 

We do not include the complete results of our experiments here due to space constraints but include them along with charts and program code on our project website. Based on this exploration we find that the best weights for this particular dataset are \emph{1} for non-propaganda and \emph{4} for propaganda and we use this to train the final model used to submit results to the leaderboard. We also found that adding Part of Speech tags and Named Entity information to BERT embeddings by concatenating these one-hot vectors to the BERT embeddings does not improve model performance. We describe these results in Section \ref{section:results}. 

\subsection{Fragment-level classification (FLC)} \label{section:flcmethod}

In addition to participating in the Sentence Level Classification task we also participate in the Fragment Level Classification task. We note that extracting fragments that are propagandistic is similar to the task of Named Entity Recognition, in that they are both span extraction tasks, and so use a BERT based model designed for this task - We build on the work by \citet{DBLP:journals/corr/abs-1906-09978} which makes use of Continuous Random Field stacked on top of an LSTM to predict spans. This architecture is standard amongst state of the art models that perform span identification. 

While the same span of text cannot have multiple named entity labels, it can have different propaganda labels. We get around this problem by picking one of the labels at random. Additionally, so as to speed up training, we only train our model on those sentences that contain some propagandistic fragment. In hindsight, we note that both these decisions were not ideal and discuss what we might have otherwise done in Section \ref{section:conclusion}.

\section{Results} \label{section:results}

In this section, we show our rankings on the leaderboard on the test set. Unlike the previous exploratory sections, in which we trained our model on part of the training set, we train models described in this section on the complete training set. 

\subsection{Results on the SLC task}

Our best performing model, selected on the basis of a systematic analysis of the relationship between cost weights and recall, places us second amongst the 25 teams that submitted their results on this task. We present our score on the test set alongside those of comparable teams in Table \ref{table:sts-results}. We note that the task description paper~\cite{EMNLP19DaSanMartino} describes a method of achieving an F1 score of 60.98\% on a similar task although this reported score is not directly comparable to the results on this task because of the differences in testing sets.

\begin{table}[ht]
\begin{center}
\footnotesize
\begin{tabular}{|C{.75cm}L{1.3cm}C{1.2cm}C{1.2cm}C{1.2cm}|}
\hline
{\footnotesize \textbf{Rank}} &    {\footnotesize \textbf{Team}}   &  {\tiny \textbf{F1} }& {\tiny \textbf{Precision} } &  {\tiny \textbf{Recall}} \\[0.5ex]
\hline
1               &     ltuorp        &   0.632375            &       0.602885            &   0.664899    \\[0.5ex]
\textbf{2}      &     \textbf{ProperGander}  &   \textbf{0.625651}            &       \textbf{0.564957}            &   \textbf{0.700954}    \\[0.5ex]
3               &     YMJA        &   0.624934            &       0.625265            &   0.624602    \\[0.5ex]

\dots               &           &              &                 &    \\[0.5ex]
20              &     Baseline      &   0.434701            &       0.388010            &   0.494168    \\[0.5ex]
\hline  
\end{tabular}
\caption{\label{table:sts-results}Our results on the SLC task (2\textsuperscript{nd}, in \textbf{bold}) alongside comparable results from the competition leaderboard.}
\end{center}
\vspace{-4mm}
\end{table}

\subsection{Results on the FLC task}

We train the model described in Section \ref{section:flcmethod} on the complete training set before submitting to the leaderboard. Our best performing model was placed 7\textsuperscript{th} amongst the 13 teams that submitted results for this task. We present our score on the test set alongside those of comparable teams in Table \ref{table:flc-results}. We note that the task description paper~\cite{EMNLP19DaSanMartino} describes a method of achieving an F1 score of 22.58\% on a similar task although, this reported score is not directly comparable to the results on this task.

\begin{table}[ht]
\begin{center}
\footnotesize
\begin{tabular}{|C{.75cm}L{1.3cm}C{1.2cm}C{1.2cm}C{1.2cm}|}
\hline
{\footnotesize \textbf{Rank}} &    {\footnotesize \textbf{Team}}   &  {\tiny \textbf{F1} }& {\tiny \textbf{Precision} } &  {\tiny \textbf{Recall}} \\[0.5ex]
\hline
1 &	newspeak &	0.248849 &	0.286299 &	0.220063    \\[0.5ex]
2 &	Antiganda &	0.226745 &	0.288213 &	0.186887    \\[0.5ex]
\dots               &           &              &                 &    \\[0.5ex]
6 &	aschern	&   0.109060 &  0.071528 &	0.229464    \\[0.5ex]
\textbf{7}  &	\textbf{ProperGander} &	\textbf{0.098969} &	\textbf{0.065167} &	\textbf{0.205634}    \\[0.5ex]
\dots               &           &              &                 &    \\[0.5ex]
11 &	Baseline &	0.000015 &	0.011628 &	0.000008    \\[0.5ex]
\hline  
\end{tabular}
\caption{\label{table:flc-results}Our results on the FLC task (7\textsuperscript{th}, in \textbf{bold}) alongside those of better performing teams  from the competition leaderboard.}
\end{center}
\vspace{-4mm}
\end{table}

One of the major setbacks to our method for identifying sentence fragments was the loss of training data as a result of randomly picking one label when the same fragment had multiple labels. This could have been avoided by training different models for each label and simply concatenating the results. Additionally, training on all sentences, including those that did not contain any fragments labelled as propagandistic would have likely improved our model performance. We intend to perform these experiments as part of our ongoing research. 

\section{Issues of Decontextualization in Automated Propaganda Detection}

It is worth reflecting on the nature of the shared task dataset (PTC corpus) and its structural correspondence (or lack thereof) to some of the definitions of propaganda mentioned in the introduction. First, propaganda is a {\it social phenomenon} and takes place as an act of communication  \citep[13-14]{oshaughnessy_politics_2005}, and so it is more than a simple information-theoretic {\it message} of zeros and ones---it also incorporates an addresser and addressee(s), each in phatic contact (typically via  broadcast media), ideally with a shared denotational code and contextual surround(s) \cite{jakobson_closing_1960}. 

As such, a dataset of decontextualised documents with labelled sentences, devoid of authorial or publisher metadata, has taken us at some remove from even a simple everyday definition of propaganda. Our models for this shared task cannot easily incorporate information about the addresser or addressee; are left to assume a shared denotational code between author and reader (one perhaps simulated with the use of pre-trained word embeddings); and they are unaware of when or where the act(s) of propagandistic communication took place. This slipperiness is illustrated in our example document (Fig. \ref{fig:kavanaugh}): note that while Sentences 3 and 7, labelled as propaganda, reflect a propagandistic attitude on the part of the journalist and/or publisher, Sentence 4---also labelled as propaganda in the training data---instead reflects a ``flag-waving" propagandistic attitude on the part of U.S. congressman Jeff Flake, via the conventions of {\it reported speech} \citep[115-130]{volosinov_marxism_1973}. While reported speech often is signaled by specific morphosyntactic patterns (e.g. the use of double-quotes and ``Flake said") \cite{spronck_reported_2019}, we argue that human readers routinely distinguish propagandistic reportage from the propagandastic speech acts of its subjects, and to conflate these categories in a propaganda detection corpus may contribute to the occurrence of false positives/negatives.

\section{Conclusions and Future Work} \label{section:conclusion}
In this work we have presented a method of incorporating cost-sensitivity into BERT to allow for better generalisation and additionally, we provide a simple measure of corpus similarity to determine when this method is likely to be useful. We intend to extend our analysis of the ability to generalise models to less similar data by experimenting on other datasets and models. We hope that the release of program code and documentation will allow the research community to help in this experimentation while exploiting these methods.

\section*{Acknowledgements}

We would like to thank Dr Leandro Minku from the University of Birmingham for his insights into and help with the statistical analysis presented in this paper.

This work was also partially supported by The Alan Turing Institute under the EPSRC grant EP/N510129/1.
Work by Elena Kochkina was partially supported 
by the Leverhulme Trust through the Bridges Programme and Warwick CDT for Urban Science \& Progress under the EPSRC Grant Number EP/L016400/1. 
\bibliography{emnlp-ijcnlp-2019}
\bibliographystyle{acl_natbib}

\end{document}